# Use of PSO in Parameter Estimation of Robot Dynamics; Part One: No Need for Parameterization

Hossein Jahandideh, Mehrzad Namvar

*Abstract*— Offline procedures for estimating parameters of robot dynamics are practically based on the parameterized inverse dynamic model. In this paper, we present a novel approach to parameter estimation of robot dynamics which removes the necessity of parameterization (i.e. finding the minimum number of parameters from which the dynamics can be calculated through a linear model with respect to these parameters). This offline approach is based on a simple and powerful swarm intelligence tool: the particle swarm optimization (PSO). In this paper, we discuss and validate the method through simulated experiments. In "Part Two" we analyze our method in terms of robustness and compare it to robust analytical methods of estimation.

## I. INTRODUCTION

The inertial and friction parameters of a robot are normally needed for exact computed torque control of the robot. There is a wealth of literature on offline methods of estimating the parameters of robots' dynamical model (e.g. [1-4]). However, most existing research on this subject is based on parameterization.

Examples of methods for obtaining the dynamics of a robot can be found in [5]. The basic formulation of a robot's inverse dynamics has the following form [5]:

$$\tau = D(q)\ddot{q} + C(q,\dot{q})\dot{q} + g(q) + F_c sign(\dot{q}) + F_v \dot{q} \qquad (1)$$

where $\tau$ denotes the vector of forces/torques applied to the robot's joints, $D$ is the manipulator inertia matrix, $C$ is the coriolis/centripetal matrix, $g$ is the gravity vector, $F_c$ is the coulomb friction and $F_v$ is the viscous friction. The vector q contains all configuration variables (displacement for prismatic joints and joint angles for revolute joints). Parameterization refers to the linear factorization of (1) as:

$$\tau = Y(q,\dot{q},\ddot{q})\alpha \qquad (2)$$

where $Y$ is the linear regressor and $\alpha$ is the set of base parameters. The base parameters are the minimum number of parameters that influence the dynamic behavior of the robot. They may be combinations of the mass, inertia, friction, and gravity parameters. An efficient procedure for reaching this factorization is clearly explained in [6].

Such parameterization makes it possible for analytical methods to estimate the parameters. The Least Squares method [2, 4], for example, which is the most widely used estimation method in the literature, depends on linear parameterization. The goal of this paper is to present a method which can estimate the values of the actual physical parameters (i.e. center-of-mass, inertia parameters, etc.) rather than the combinations of them, without getting involved with the parameterization procedure. This method is based on the particle swarm optimization (PSO).

Swarm and evolutionary algorithms are optimization tools that are inspired by natural phenomena. Examples of swarm and evolutionary algorithms and their various applications have been discussed in an orderly manner in [7]. The PSO, in particular, was motivated by the simulation of bird flocking or fish schooling. PSO was first introduced by Kennedy and Eberhart in [8], and analyzed thoroughly by Clerk and Kennedy in [9]. As an optimization technique, PSO found way into numerous applications throughout the years. Examples of the applications of PSO in robotics are discussed in [3, 7].

PSO quickly found way into system identification problems. [10], for example, presented a method based on artificial neural networks and PSO for the identification of a general dynamical system. However, in [10] and similar papers, identification of a system requires building a model to predict the input-output behavior of the system, while in robots, the exact form of the dynamical model is obtained from the kinematics of the robot according to [5]. If the kinematics is uncertain, there are methods to estimate the kinematics; [11] for example, presented a PSO approach to the kinematics estimation. This being said, the exact form of a robot's dynamical model is known, and strong analytical methods exist that can estimate the values of the parameters in the model. Hence, PSO has rarely been used in estimating the dynamical model parameters of a robot.

However, [3] applies PSO to estimating the dynamical model parameters of the first three links of a Staubli RX-60 Robot and compares the method to the conventional least squares method. In [3], parameterization is used as a preliminary step to estimating the parameters through PSO. In this paper, this preliminary step (parameterization) is removed, and simulation results are shown to support our method. We have simulated the estimation of all parameters of the three links of a cylindrical robot to demonstrate the simplicity and efficiency of our approach.

It is important to note that all offline methods of dynamic model parameter estimation utilize experimental samples, thus the way we obtain these samples (i.e. the way we excite our system to obtain these samples) plays an important role in the accuracy and reliability of our estimation. In this paper, we have used PSO not only in the estimation procedure, but also in our trajectory planning.

Mehrzad Namvar is a faculty member of the control systems group in the Electrical Engineering department at Sharif University of Technology, Tehran, Iran. He received PhD in Control systems in 2001 from the Grenoble Institute of Technology (INPG) in France.  namvar@sharif.ir

Hossein Jahandideh is a student in the Electrical Engineering department at Sharif University of Technology, Tehran, Iran (2008-2013) hs.jahan@gmail.com

The rest of this paper is arranged as follows: In section II, the PSO algorithm is explained. Section III describes how PSO can be used without the preliminary step of parameterization and section IV notes the advantages of using this method. Section V introduces the cylindrical robot and the parameters to be estimated. Section VI explains the application of PSO in planning an acceptable excitation trajectory. Section VII is dedicated to a simulated experiment. Finally, section VIII concludes the paper.

## II. Particle Swarm Optimization

Let $f: \Re^n \rightarrow \Re$ be the function to be optimized. Without loss of generality, we'll take our objective to be minimization.

Objective:    minimize $f(x)$

The PSO algorithm assigns a swarm of k particles to search for the optimal solution in an n-dimensional space. The starting position of a particle is randomly set within the range of possible solutions to the problem. The range is determined based on an intuitive guess of the maximum and minimum possible values of each component of x, but doesn't need be accurate. Each particle analyzes the function value ($f(p)$) of its current position (p), and has a memory of its own best experience (Pbest), which is compared to p in each iteration, and is replaced by p if $f(p)<f(Pbest)$. Besides its own best experience, each particle has knowledge of the best experience achieved by the entire swarm (the global best experience denoted by Gbest). Based on the data each agent has, its movement in the i-th iteration is determined by the following formula:

$$V_i = w_i V_{i-1} + C_1 r_1 (Pbest - p_{i-1}) + C_2 r_2 (Gbest - p_{i-1}) \quad (3)$$

where $V_i$, $P_i$, Pbest, and Gbest are n-vectors (or similar objects, such as matrices with n components), $r_1$ and $r_2$ are random numbers between 0 and 1, re-generated at each iteration. $C_1$ and $C_2$ are constant positive numbers, $C_1$ is the *cognitive learning rate* and $C_2$ is the *social learning rate*. $w_i$ is the *inertia weight*. The new position of each particle at the i-th iteration is updated by:

$$p_i = p_{i-1} + V_i \quad (4)$$

After certain conditions are met, the iterations stop and the Gbest at the latest iteration is taken as the optimal solution to the problem. In this paper, we let the PSO algorithm end when the number of iterations reaches a certain number

## III. PSO in Parameter Estimation of Robot Dynamics

Each sample we have of the robot dynamics contains the following data:

$\tau_{(i)}$, $q_{(i)}$, $\dot{q}_{(i)}$, $\ddot{q}_{(i)}$,    where $\tau$ is the n-vector of forces/torques, and q is the state of the n joint variables (n is equal to the degrees of freedom of the robot). The index (i) is used for the i-th sample. For each sample, based on the estimated parameters and the inverse dynamics model, a vector $\hat{\tau}$ can be calculated. Referring to (1), we have:

$$\hat{\tau}_{(i)} = D_{est}(q_{(i)})\ddot{q}_{(i)} + C_{est}(q_{(i)},\dot{q}_{(i)})\dot{q}_{(i)} + g(q_{(i)}) + F_{c_{est}}sign(\dot{q}_{(i)}) + F_{v_{est}}\dot{q}_{(i)} \quad (5)$$

where $D_{est}$, $C_{est}$, $F_{vest}$, and $F_{cest}$ are calculated according to the estimated parameters. Thus, for every set of potential estimates of the parameters, we have a $\tau$ and a $\hat{\tau}$. Define $e_{(i)}$ for the i-th sample:

$$e_{(i)} = \tau_{(i)} - \hat{\tau}_{(i)} \quad (6)$$

Now define the matrix e for which the i-th column is $e_{(i)}$:

$$E = [e_{(1)} | e_{(2)} | ... | e_{(N)}] \quad (7)$$

where N is the number of samples available. The cost function for the PSO algorithm is defined for the matrix *E*. An analysis of what function to define in this stage is presented in "Part Two". In this paper we define the cost function as:

$$f_{(e)} = \|E\|_2 \quad (8)$$

The objective of the PSO algorithm in our estimation task is to find the set of parameters that minimizes the cost function $f$.

Software for simulating robot dynamics (both symbolic and numeric) are introduced in [12]. The software used in our simulation, is *Robotics Toolbox for Matlab* [13]. This software, just as many other robotics software, can numerically and efficiently calculate the inverse dynamics of a robot; meaning that given the kinematics of the robot links, and a set of potential estimates for the mass, center of mass, inertia, and frictional parameters, a model of the robot is simulated which can perform the inverse dynamics function of the robot. Thus for every set of potential estimates of the parameters, a cost function can be calculated via (6-8). If the position of each particle in the PSO swarm is defined as an estimate of the robot parameters, the PSO algorithm can find the estimate which minimizes (8). The algorithm was programmed in Matlab software: For each particle, the toolbox simulates the inverse dynamics of the robot and uses it on all samples to calculate $\hat{\tau}_{(i)}$; the function value of the particle is then calculated through (6-8); Pbest and Gbest are updated and the next move is obtained from (3). As can be seen, this method does not require the parameterization step.

## IV. Advantages

When the need for parameterization is omitted, each particle of the PSO swarm can directly represent the mass, center of mass, inertia, and frictional parameters, rather than combinations of them. This way, a physical insight is given of the characteristics of the robot.

When parameterization is required, the parameters become very complicated as the degrees of freedom increase. Even if we use software such as *ScrewCalculus*

[14] to accomplish the symbolic parameterization task, the user must organize the symbolic parameters to prepare them for estimation via numeric algorithms, which can be a frustrating task, considering the complicated symbolic analysis.

To estimate the components of α (defined in (2)) via a conventional method of estimation, the matrix Y must be encoded. For more complicated robots and higher degrees of freedom, Y and α become too complicated for analysis.

After the estimation procedure, the robot model is ready to be used for control purposes, for either model based or non-model based control methods.

For control methods which require only the prediction of the input-output behavior of the model, a robot can be simulated in the same software used for estimation, which simulates the input-output behavior of the robot.

The identification procedure results in a simulation model of the robot from which the *D, C,* and *g* matrices defined in (2) can also be calculated by the robot simulation software, given the state of the system. Thus, control methods which require calculation of the mentioned matrices can also be implemented after this simple parameter estimation procedure is carried out.

## V. THE CYLINDRICAL ROBOT

We have used the cylindrical robot in our simulations for simplicity of presentation. The link parameters of a cylindrical robot are shown in table 1 [4].

TABLE I.  THE LINK PARAMETERS OF THE CYLINDRICAL ROBOT

| link number (i) | $a_{(i)}(m)$ | $\alpha_{(i)}(rad)$ | $d_{(i)}(m)$ | $q_{(i)}$ |
|---|---|---|---|---|
| 1 | 0 | 0 | 0 | $\theta_1$ |
| 2 | 0 | $-\pi/2$ | 0 | $d_2$ |
| 3 | 0 | 0 | 0 | $d_3$ |

The parameters that constitute the robot dynamics are as follows: $s_{ia}$ is the component of the center of mass of the i-th link along its own a-axis (a could be x, y, or z); $m_i$ is the mass of the i-th link, and $I_{iab}$ is the ab component of the moment of inertia of the i-th link about its center of mass. $f_c$ and $f_v$ are the coulomb and viscous frictions respectively.

If the links are treated as one dimensional figures and the frictional factors are omitted (as they are in [4], and in section VI of this paper), this robot has only 4 identifiable parameters ($m_2$, $m_3$, $s_{3z}$, and $I_{1zz}+I_{2yy}+I_{3yy}$). For a more realistic simulation, such assumptions have not been made in section VII of this paper.

## VI. EXCITATION TRAJECTORY

Defining a proper trajectory by which to excite a dynamical system for sampling purposes, plays an important role in the accuracy of the estimation based on the obtained samples. If part of a system is not excited, the parameters pertaining to the dynamics of that part will not be identifiable. Similarly, if part of a system is insufficiently excited, small errors on the measurement samples can cause large errors in the estimated values of the parameters pertaining to the insufficiently excited system variables. The excitation trajectory must be planned in a way that the physical constraints on the system are observed (if they are not, the planned trajectory will not be executable) and all state variables are properly excited.

Much research has been carried out on planning excitation trajectories for robots. Particularly, [15] uses the genetic algorithm to obtain an optimal trajectory for excitation. In [15], it is mathematically proven that the larger the determinant of the square matrix $W^T R_v^{-1} W$, the less is the error of our least squares estimation; where $R_v$ is the autocorrelation matrix of the error vector $\tau_{sam}-\tau_{est}$, and $W$ is built by combining all the $Y_{(i)}$ obtained from measurement samples ($Y$ is the regressor defined in (2)).

$$W = \begin{bmatrix} Y_{(q1,\dot{q}1,\ddot{q}1)} \\ Y_{(q2,\dot{q}2,\ddot{q}2)} \\ ... \\ Y_{(qN,\dot{q}N,\ddot{q}N)} \end{bmatrix} \qquad (9)$$

When $R_v$ is unknown, it can be replaced by the identity matrix, thus the objective of the genetic algorithm becomes maximizing the determinant of the square matrix $W^T W$.

$$H = |W^T W| \qquad (10)$$

Objective:  Maximize H
Subject to physical constraints on the joint variables

A perceivable explanation of this theorem is that in order to maximize H, the components of the regressor must be large as well as properly varied, thus the system states are well excited.

Thus, a cognitive hypothesis is that if $W$ is replaced by $Q_{sam}$ as defined below, the obtained trajectory will remain an efficient excitation. If the hypothesis is correct, we can obtain an efficient trajectory without parameterization.

$$Q_{sam} = \begin{bmatrix} q_{11}|\dot{q}_{11}|\ddot{q}_{11}|q_{12}|\dot{q}_{12}|\ddot{q}_{12}|...|q_{1n}|\dot{q}_{1n}|\ddot{q}_{1n} \\ q_{21}|\dot{q}_{21}|\ddot{q}_{21}|q_{22}|\dot{q}_{22}|\ddot{q}_{22}|...|q_{2n}|\dot{q}_{2n}|\ddot{q}_{2n} \\ ... \\ q_{N1}|\dot{q}_{N1}|\ddot{q}_{N1}|q_{N2}|\dot{q}_{N2}|\ddot{q}_{N2}|...|\dot{q}_{Nn}|\ddot{q}_{Nn} \end{bmatrix} \qquad (11)$$

where $q_{ij}$ is the j-th joint variable of the i-th sample.

We replace the genetic algorithm by PSO and must follow the following procedure for obtaining the desired trajectory:

1- Define:

$q_j = a_{1j}\sin(\omega_{1j}t) + a_{2j}\sin(\omega_{2j}t) + a_{3j}\sin(\omega_{3j}t) \quad j=1,2,...,n$ (12)

(n is the degrees of freedom of the robot.)

2- Define $a_{kj}$ and $\omega_{kj}$ as the parameters determined by PSO. (6n variables in total)

3- Calculate the samples (defined by (12)) at the times $\frac{T}{N}i$ ($i = 1,...,N$) and place them inside $Q_{sam}$.

(T is the sampling time and N is the number of desires samples.)

4- Define :

$$H_Q = abs(|Q_{sam}^T Q_{sam}|) - a.10^{40} \quad (13)$$

where *a* is a binary value, such that it is 0 if all samples meet the physical constraints, and is 1 if any component of any sample fails to meet the physical constraints. The function *abs* denotes the absolute value function. Note that it is possible that none of the samples break the constraints, while the trajectory does break the constraints at some point in time. Thus, we must program the constraints slightly above their lower limits and slightly below their higher limits.

5- Use the PSO algorithm to maximize $H_Q$ and determine the values of $a_{kj}$ and $\omega_{kj}$. (k=1,2,3 j=1,2,…,n)

We will test our hypothesis through a simulated experiment. Consider the example of the cylindrical robot. Consider the following constraints to simplify the example:

$$\begin{array}{l} 0 \leq q_j \leq 1 \\ -1 \leq \dot{q}_j, \ddot{q}_j \leq 1 \end{array} \quad (j = 1, 2, 3) \quad (14)$$

The starting point for the joint variables is considered to be: $(q_1, q_2, q_3) = (0.5, 0.5, 0.5)$

The PSO parameters of the trajectory planned by the above procedure and two random trajectories (which meet the constraints) are given in table 2. These trajectories were used as reference trajectories to obtain measurement samples for each. The three sets of samples were affected by the same noise disturbance. The PSO algorithm for parameter estimation of the robot's dynamics was run 5 times for each set of samples and the results are compared in table 3.

It is seen in table 3 that the estimation derived from the samples of the obtained trajectory are very close to the real values, even though the samples had errors. The algorithm reached the exact same estimates in all its 5 runs.

For the random trajectories, we see a large variance in the estimated values, and the perturbations have caused relatively large errors on the average estimated values. Our hypothesis is validated by this simulation. This method of trajectory planning is used in the next section of this paper and throughout "part two".

TABLE II. THE PLANNED AND RANDOM TRAJECTORY PARAMETERS

|  | PSO planned trajectory | Random trajectory 1 | Random trajectory 2 |
|---|---|---|---|
| $a_{11}$ | 0.0184 | 0.0494 | 0.0061 |
| $\omega_{11}$ | 0.0113 | 0.1000 | 0.0869 |
| $a_{21}$ | 0.2105 | 0.0085 | 0.0440 |
| $\omega_{21}$ | 2.2918 | 0.0688 | -0.7861 |
| $a_{31}$ | 0.3841 | 0.0486 | 0.0973 |
| $\omega_{31}$ | 0.5601 | 0.0332 | -0.0211 |
| $a_{12}$ | 0.0308 | -0.0076 | 0.1192 |
| $\omega_{12}$ | 0.0624 | 0.0771 | 0.1507 |
| $a_{22}$ | -0.2101 | 0.0546 | 0.1028 |
| $\omega_{22}$ | -0.0124 | 0.0709 | 0.0928 |
| $a_{32}$ | 0.5056 | 0.0614 | -0.1604 |
| $\omega_{32}$ | 1.3691 | -0.1922 | 0.0903 |
| $a_{13}$ | 0.3410 | 0.0590 | 0.1019 |
| $\omega_{13}$ | 1.8234 | 0.0486 | 0.1026 |
| $a_{23}$ | -0.1658 | 0.0082 | 0.0765 |
| $\omega_{23}$ | -0.3325 | 0.0563 | 0.0726 |
| $a_{33}$ | 0.2849 | 0.0553 | 0.0726 |
| $\omega_{33}$ | -0.8144 | 0.0309 | 0.0349 |

TABLE III. ESTIMATED VALUES BASED ON ONE PLANNED TRAJECTORY AND TWO RANDOM TRAJECTORIES

|  | Planned | Random 1 | Random 2 | Real values |
|---|---|---|---|---|
| average $m_2$ | 5.003 | 4.09 | 4.72 | 5 |
| max-min $m_2$ | 0 | 2.71 | 1.65 | - |
| average $m_3$ | 2.996 | 3.91 | 3.28 | 3 |
| max-min $m_3$ | 0 | 2.71 | 1.64 | - |
| average $-s_{3z}$ | 0.496 | 0.62 | 0.66 | 0.5 |
| max-min $-s_{3z}$ | 0 | 1.22 | 0.64 | - |
| average $I_{1zz}+I_{2yy}+I_{3yy}$ | 3.052 | 1.50 | 2.83 | 3 |
| min-max $I_{1zz}+I_{2yy}+I_{3yy}$ | 0 | 2.21 | 0.40 | - |
| average cost function | 0.285 | 0.029 | 0.031 | - |

## VII. SIMULATION RESULTS

In order to achieve a realistic simulation, all physical parameters have been considered and estimated. The trajectory suggested by the PSO algorithm (16) is taken as the reference trajectory in our sampling operation. All sampling data have been given a random error of up to 10% of their total value. The PSO parameters used in our simulation and the average computational time for each run are summarized in table 4. The algorithm was run 10 times; the results are shown in tables 5-9, and the unidentifiable parameters have been disclosed. A parameter is classified as unidentifiable if there is too much variation in its estimated value. Note that all real values of the robot parameters and the physical constraints are dictated to the simulated robot; the simulated robot in our experiment is not meant to model a specific counterpart in the real world.

TABLE IV. PSO PARAMETERS USED IN SIMULATION AND AVERAGE CPU TIME

| swarm population | number of iterations | $c_1, c_2$ | $w_i$ | average CPU time |
|---|---|---|---|---|
| 20 | 100 | 1.3 | 0.6 | 9.37 seconds |

The physical constraints to be observed by the trajectory planning algorithm are assumed to be as follows:

$$-\pi \leq \theta_1(rad) \leq \pi, \quad -4 \leq \dot{\theta}_1(\tfrac{rad}{s}) \leq 4, \quad -3 \leq \ddot{\theta}_1(\tfrac{rad}{s}) \leq 3$$
$$0 \leq d_2(m) \leq 1, \quad -2 \leq \dot{d}_2(\tfrac{m}{s}) \leq 2, \quad -2 \leq \ddot{d}_2(\tfrac{m}{s}) \leq 2 \quad (15)$$
$$0 \leq d_3(m) \leq 1, \quad -1.5 \leq \dot{d}_3(\tfrac{m}{s}) \leq 1.5, \quad -1 \leq \ddot{d}_3(\tfrac{m}{s^2}) \leq 1$$

The following trajectory was planned for the experiment:

$$\theta_{1(t)} = 0.97\sin(1.15t) + 0.83\sin(1.1t) + 1.94\sin(0.42t) - 2.63$$
$$d_{2(t)} = 0.96\sin(0.57t) + 0.35\sin(2.05t) - 1.1\sin(0.12t) + 0.11$$
$$d_{3(t)} = -2.3\sin(0.07t) + 0.32\sin(1.5t) + 1.42\sin(0.38t) - 0.08$$
$$0 \leq t \leq 10$$
(16)

It is important to note that once the unidentifiable parameters are recognized, in simulating the robot for control or similar purposes, we are not free to set the values of the unidentifiable parameters to arbitrary values. As an example, in the cylindrical robot, $I_{1zz}$, $I_{2yy}$, and $I_{3yy}$, all are unidentifiable parameters. The summation of these parameters, however, is crucial to the robot dynamics. We will define such parameters as *semi-identifiable* (SI). Some parameters, such as $m_1$ in this example, are not completely absent in a robot's dynamics, but have such little effect that renders it unidentifiable. We will define such parameters as *nearly unidentifiable* (NUI). For simulation purposes after parameter estimation, all SI and NUI parameters must be set in accordance with the suggestion of the PSO algorithm. In tables 5-7, SI or NUI parameters have been recognized by the following procedure:

A parameter classified as not identifiable by the PSO database is classified as SI/NUI if the total cost function is affected by the variation of that specific parameter while all other parameters are kept unchanged; otherwise the parameter is classified as unidentifiable. It is difficult to identify whether a parameter is SI or NUI, but it can usually be said that a relative variation in the estimated values for the SI parameters cause larger variation in the cost function than does the same relative variation in the estimated value of an NUI parameter.

For further verification, the mean estimated values (including of the SI, NUI, and UI parameters) have been used to simulate a second robot (aside from the simulation representing the real robot) and the following reference trajectory was given to both robots to compare the resulting force/torques and the results are shown in figures 1 to 3:

$$\theta_{1(t)} = 1.97\sin(0.5t) + 0.44\sin(2.2t) + 0.35\sin(0.9t) - 2.7$$
$$d_{2(t)} = 0.6\sin(1.7t) - 0.3\sin(1.45t) + 0.86\sin(0.7t) - 0.06$$
$$d_{3(t)} = 0.4\sin(0.3t) + 0.4\sin(1.3t) + 0.13\sin(1.2t) + 0.16$$
$$0 \leq t \leq 10$$
(17)

TABLE V. SIMULATION RESULTS FOR LINK 1

| parameter | real value | mean estimate | Coefficient of Variation | max-min estimate | status (I, UI, SI) |
|---|---|---|---|---|---|
| M | 2 | 1.51 | 0.64 | 3.27 | NUI |
| -$s_x$ | 0.5 | 0.48 | 0.48 | 0.69 | NUI |
| -$s_y$ | 0.5 | 0.39 | 0.52 | 0.7 | NUI |
| -$s_z$ | 1 | 0.52 | 0.79 | 1.08 | UI |
| $I_{xx}$ | 4 | 2.97 | 0.61 | 5.96 | UI |
| $I_{yy}$ | 1 | 0.87 | 0.50 | 1.72 | UI |
| $I_{zz}$ | 4 | 3.23 | 0.49 | 5.36 | SI |
| $I_{xy}$ | 1 | 0.84 | 0.41 | 1.53 | UI |
| $I_{yz}$ | 1 | 0.89 | 0.69 | 2.33 | UI |
| $I_{xz}$ | 1 | 0.96 | 1.11 | 3.66 | UI |
| $f_c$ | 1 | 1.09 | 0.34 | 1.72 | SI/NUI |
| $f_v$ | 1 | 0.96 | 0.36 | 1.5 | SI/NUI |

TABLE VI. SIMULATION RESULTS FOR LINK 2

| parameter | real value | mean estimate | Coefficient of Variation | max-min estimate | status (I, UI, SI) |
|---|---|---|---|---|---|
| M | 5 | 4.70 | 0.08 | 1.21 | I |
| -$s_x$ | 0.5 | 0.75 | 0.40 | 1.06 | NUI |
| -$s_y$ | 0.5 | 0.51 | 0.48 | 0.80 | NUI |
| -$s_z$ | 0.5 | 0.35 | 0.70 | 0.76 | NUI |
| $I_{xx}$ | 3 | 2.23 | 0.29 | 2.38 | UI |
| $I_{yy}$ | 1 | 0.79 | 0.66 | 1.74 | SI |
| $I_{zz}$ | 3 | 2.42 | 0.44 | 3.19 | UI |
| $I_{xy}$ | 1 | 0.83 | 0.49 | 1.25 | UI |
| $I_{yz}$ | 1 | 0.98 | 0.86 | 2.19 | UI |
| $I_{xz}$ | 1 | 1.38 | 1.33 | 5.99 | UI |
| $f_c$ | 1 | 0.996 | 0.34 | 1.00 | SI/NUI |
| $f_v$ | 1 | 1.006 | 0.64 | 1.90 | SI/NUI |

TABLE VII. SIMULATION RESULTS FOR LINK 3

| parameter | real value | mean estimate | Coefficient of Variation | max-min estimate | status (I, UI, SI) |
|---|---|---|---|---|---|
| M | 3 | 2.98 | 0.03 | 0.25 | I |
| -$s_x$ | 0.5 | 0.54 | 0.04 | 0.067 | I |
| -$s_y$ | 0.5 | 0.38 | 0.65 | 0.90 | NUI |
| -$s_z$ | 0.5 | 0.51 | 0.007 | 0.014 | I |
| $I_{xx}$ | 2 | 1.69 | 0.50 | 3.24 | UI |
| $I_{yy}$ | 2 | 1.84 | 0.47 | 2.43 | SI |
| $I_{zz}$ | 2 | 0.80 | 0.49 | 1.32 | UI |
| $I_{xy}$ | 0.5 | 0.53 | 0.98 | 1.78 | UI |
| $I_{yz}$ | 0.5 | 0.40 | 0.41 | 0.44 | UI |
| $I_{xz}$ | 0.5 | 0.34 | 0.65 | 0.74 | UI |
| $f_c$ | 1 | 0.77 | 0.37 | 0.89 | SI/NUI |
| $f_v$ | 1 | 0.93 | 0.17 | 0.54 | I |

As seen in tables 5-7, unidentifiable parameters have been recognized and acceptable estimates of the identifiable parameters have been given. The estimated parameters are verified for being acceptable by the result of a verification simulation (figures 1 to 3). The same algorithm may be used to estimate the parameters of any industrial robot, given the robot's link parameters and samples obtained from an acceptable (though not necessarily perfect) excitation trajectory. The performance of the estimated robot is verified by figures 1 to 3. It is seen that even though some of the mean estimations in tables 5-7 differ greatly from their real values, the behavior of the estimated robot follows closely the behavior of the real robot. This confirms that the parameters classified as SI/NUI or UI have been correctly discovered.

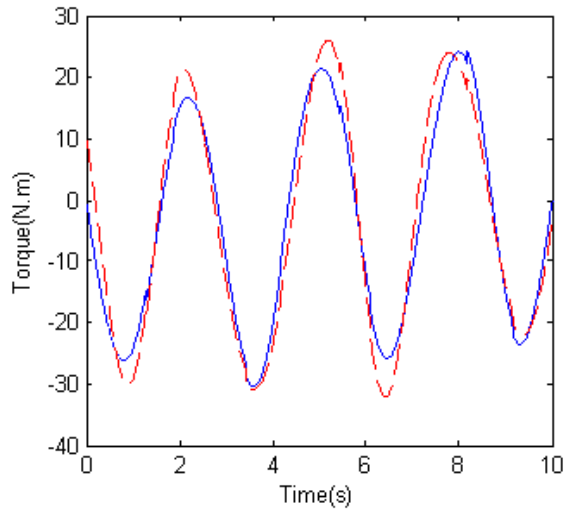

Figure 1. Torque comparison for the verification trajectory; Joint 1. solid line: real robot; dashed line: estimated robot

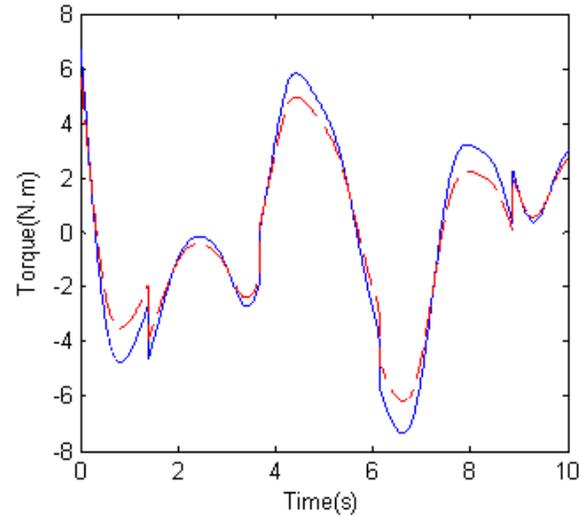

Figure 3. Torque comparison for the verification trajectory; Joint 3. solid line: real robot; dashed line: estimated robot

## VIII. CONCLUSION

In this paper, a time-efficient, cost-effective, easily implemented, and flexible method based on particle swarm optimization was applied to the parameter estimation of robot dynamics. As shown through simulation on a cylindrical robot, this method is easily executable on any industrial robot with any number of degrees of freedom. With this method, any user with access to robot simulation software can identify a robot and prepare it for control or other purposes without getting involved with parameterization. In order to completely avoid parameterization, the excitation trajectory was also planned based on a PSO approach which requires only the link parameters and the physical constraints of the joint variables. In "part 2", this method of robot parameter estimation is compared to least squares, total least squares, and robust least squares methods in terms of robustness toward relatively large errors in the sample data.

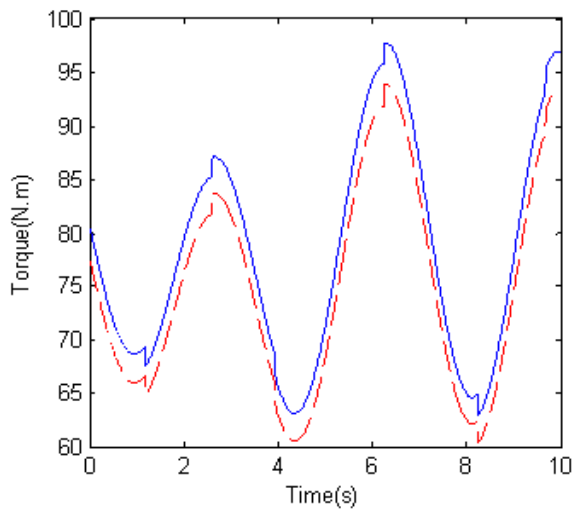

Figure 2. Torque comparison for the verification trajectory; Joint 2. solid line: real robot; dashed line: estimated robot

REFERENCES

[1] Basilio B. and Aldo C., "Identification of Industrial Robot Parameters for Advanced Model-Based Controllers Design", IEEE International Conference on Robotics and Automation, p. 1693-1698, April 2005.
[2] Chan S., "An Efficient Algorithm for Identification of Robot Parameters Including Drive Characteristics", Journal of Intelligent and Robotic Systems 32: 291–305, 2001.
[3] Bingul Z. and Karahan O., "Dynamic identification of Staubli RX-60 robot using PSO and LS methods", Expert Systems with Applications, Volume 38, 2011.
[4] Khosla P. and Kanade T., "Parameter Identification of Robot Dynamics", Proc. 24th Conf. Decision and Control, December 1985.
[5] De Luca A. and Ferrajoli L., "A Modified Newton-Euler Method for Dynamic Computations in Robot Fault Detection and Control", IEEE Intern. Conf. Robotics and Automation, Kobe, Japan, May 12-17, 2009.
[6] Gabiccini M., Bracci A., Artoni A., "Direct Derivation of the Dynamic Regressors for Serial Manipulators with Mixed Rigid/Elastic Joints" XX Congresso AIMETA, Bologna, Italy (2011).
[7] Mostajabi T., Poshtan J., "Control and System Identification via Swarm and Evolutionary Algorithms", International Journal of Scientific and Engineering Research Volume 2, Issue 10, October 2011.
[8] Kennedy J. and Eberhart R., "Particle Swarm Optimization", Proceedings of IEEE International Conference on Neural Networks, Piscataway, NJ. pp. 1942– 1948, 1995.
[9] Clerc M. and Kennedy J. "The Particle Swarm: Explosion, Stability, and Convergence in a Multi-Dimensional Complex Space", IEEE Transactions on Evolution Computer, 6(1): 58-73, 2002.
[10] Omkar S. and Mudigere D. "Non-Linear Dynamical System Identification Using Particle Swarm Optimization", Proc. Int. Conf. Advances in Control and Optimization of Dynamical Systems, 2007.
[11] Barati M., Khoogar A., and Nasirian M., "Estimation and Calibration of Robot Link Parameters with Intelligent Techniques", Iranian J. Electrical and Electronic Eng., Volume 7, No. 4, Dec. 2011.
[12] Toz M., Kucuk S., "Dynamics Simulation Toolbox for Industrial Robot Manipulators", Computer Applications in Engineering Education, Volume 18, Issue 2, pages 319–330, June 2010.
[13] Corke P., "Robotics TOOLBOX for MATLAB", CSIRO Manufacturing Science and Technology, 2001.
[14] Gabiccini M., "ScrewCalculus: a Mathematica Package for Robotics," DIMNP, University of Pisa, http://www2.ing.unipi.it/d11181/RAR/ScrewCalculus.rar, 2009.
[15] Calafiore G., Indri M., and Bona B., "Robot Dynamic Calibration: Optimal Excitation Trajectories and Experimental Parameter Estimation", J. Robotic Systems, Vol. 18, Issue 2, pp. 55–68, Feb. 2001.